\documentclass[10pt,twocolumn,letterpaper]{article}

\usepackage{iccv}
\usepackage{times}
\usepackage{epsfig}
\usepackage{graphicx}
\usepackage{amsmath}
\usepackage{amssymb}
\usepackage{bm}
\usepackage{soul}
\usepackage{multirow}
\usepackage{epsfig}
\usepackage{graphicx}
\usepackage{amsmath}
\usepackage{booktabs}
\usepackage{bbm}

\usepackage[breaklinks=true,bookmarks=false]{hyperref}

\iccvfinalcopy 

\def\iccvPaperID{****} 

\ificcvfinal\fi

\begin{document}

\title{\textsc{M3DETR}: Multi-representation, Multi-scale, Mutual-relation 3D Object Detection with Transformers}

\author{
Tianrui Guan$^{1}$\thanks{Equal contribution.} 
~~~Jun Wang$^{1}$\footnotemark[1]
~~~ Shiyi Lan$^{1}$\thanks{Corresponding author: \textless{}\href{mailto:sylan@cs.umd.edu}{sylan@cs.umd.edu}\textgreater{}.}
~~~ Rohan Chandra$^{1}$
~~~ Zuxuan Wu $^{2}$ \\
~~~ Larry Davis $^{1}$
~~~ Dinesh Manocha$^{1}$\\
$^1$University of Maryland, College Park ~~~~~
$^2$Fudan University\\
\tt\small \{rayguan, sylan, rohan\}@cs.umd.edu, \tt\small  \{junwang,lsd\}@umiacs.umd.edu, \\ \tt\small  zxwu@fudan.edu.cn,   \tt\small  dmanocha@umd.edu
}

\maketitle

\begin{abstract}
We present a novel architecture for 3D object detection,  \textsc{M3DETR}, which combines different point cloud representations (raw, voxels, bird-eye view) with different feature scales based on multi-scale feature pyramids. \textsc{M3DETR} is the first approach that unifies multiple point cloud representations, feature scales, as well as models mutual relationships between point clouds \textit{simultaneously} using transformers. We perform extensive ablation experiments that highlight the benefits of fusing representation and scale, and modeling the relationships. Our method achieves state-of-the-art performance on the KITTI 3D object detection dataset and Waymo Open Dataset. Results show that \textsc{M3DETR} improves the baseline significantly by 1.48\% mAP for all classes on Waymo Open Dataset.  In particular, our approach ranks $1^{st}$ on the well-known KITTI 3D Detection Benchmark for both car and cyclist classes, and ranks $1^{st}$ on Waymo Open Dataset with single frame point cloud input. Our code is available at: https://github.com/rayguan97/M3DETR.
\end{abstract}

\section{Introduction}
3D object detection is a fundamental problem in computer vision and many applications, including autonomous driving~\cite{geiger2012we, Sun_2020_CVPR}, augmented reality~\cite{park2008multiple} and robotics~\cite{oh2002development}.
Moreover, different methods have been proposed for various sensors, including monocular cameras, depth cameras, LiDAR, and radars ~\cite{qi2017pointnet,qi2017pointnet++,qi2018frustum,yang2019std,ye2020hvnet,yang20203dssd,shi2020pv,yin2020center}. 2D object detection deals with detecting objects from RGB images and videos, while 3D object detection utilizes point cloud-based representations obtained from LiDARs or other sensors. Moreover, it is known that point cloud data obtained from LiDAR sensors tends to be more accurate than RGB images and videos~\cite{caesar2020nuscenes}. Consequently, point clouds are being widely used for scene understanding in autonomous driving and AR. 

\begin{figure}[t]
    \centering
    \includegraphics[width=\linewidth]{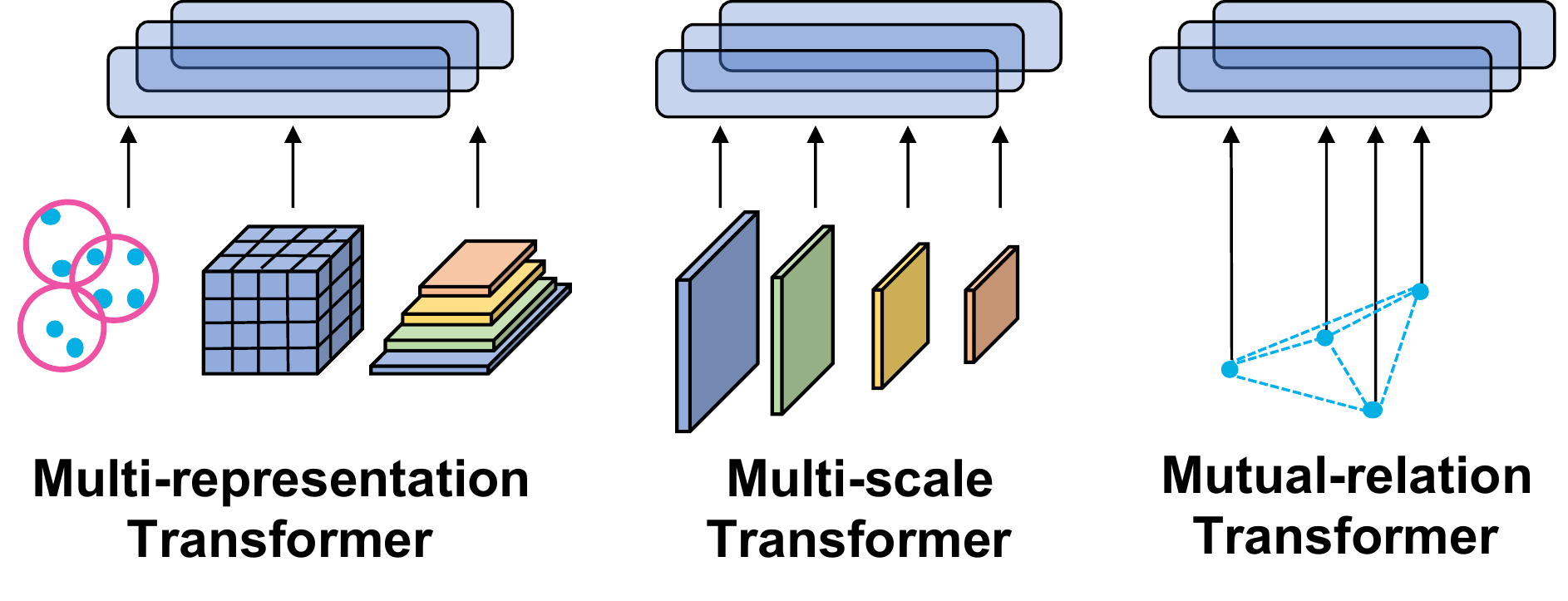}
    \caption{M3 Transformers used in our approach: Left: Multi-representation transformers; Middle: Multi-scale transformers; Right: Mutual-relation transformers. We use these characteristics to present a novel 3D object detection architecture.}
    \label{fig:coverpic.p}
\vspace{-0.5cm}
\end{figure}

The previous
state-of-the-art methods for 3D object detection base on different networks~\cite{yin2020center,shi2020pv}. However, there are two key limitations:

\noindent{\textbf{Ineffective point cloud representations:}} The three major techniques used to process point clouds are based on voxels~\cite{zhou2018voxelnet,yan2018second}, raw point clouds~\cite{qi2017pointnet,qi2017pointnet++,qi2018frustum,shi2019pointrcnn,yang2019std,liu2019point, qin2020weakly}, and bird's-eye-view ~\cite{chen2017multi, ku2018joint, yang2018pixor}. Each representation has a unique advantage and it has been shown that combining these representations can result in terms of detection accuracy \cite{qi2018frustum,shi2020pv,chen2019fast}. However, fusing these representations is non-trivial. First, the architectures corresponding to the VoxelNets, the PointNets, and the 2D convolutional neural networks are  different. Moreover, raw point clouds need to be converted to voxels and pixels before techniques based on VoxelNets and 2D convolutional neural networks can be applied. The differences between the inputs of these three neural models can result in semantic gaps. Previous works~\cite{chen2019fast,shi2020pv} tend to use feature concatenation and attention modules to fuse multi-representation features. However, the correlation between features of different representations has not been addressed.

\noindent{\textbf{Insufficient modeling of multi-scale features:}} Fusing multi-scale feature maps is a well-known technique used for improving the detection performance of 2D object detection. In terms of 3D object detection, current approaches tend to use multi-scale feature pyramids~\cite{qi2017pointnet,qi2017pointnet++,yang20203dssd,lan2019modeling,lang2019pointpillars,ye2020hvnet}. However, fusing these multiple feature pyramids is non-trivial because the higher resolution and the larger receptive fields are conflicting \cite{lin2017feature, qi2017pointnet++}. Existing methods~\cite{qi2017pointnet++,ye2020hvnet,shi2020pv} fuse the multi-scale features using bi-linear down-sampling/up-sampling and concatenation. Although these approaches~\cite{qi2017pointnet++,ye2020hvnet,shi2020pv} can improve the accuracy by a large margin, there are many challenges with respect to the underlying fusion method  in terms of the correlation between feature maps of different scales. 

A key issue in terms of designing a good approach for object detection is exploiting the correlation between different representations and the large size of the receptive fields.  Our approach is motivated by use of transformers~\cite{vaswani2017attention}, a form of neural network based on attention that has been used in natural language processing ~\cite{hu2020iterative,tan2019lxmert,su2019vl,lu2019vilbert}. Specifically, transformers use multi-head attention to narrow the semantic gap between different representations by adapting to informative features and eliminating noise. 

Another key aspect of 3D object detection is to model the mutual relationships between different points in the point cloud data~\cite{qi2017pointnet,qi2017pointnet++,lan2019modeling,thomas2019kpconv}. Modeling mutual relationships can enhance the ability to recognize the fine-grained patterns and can generalize to complex scenes.
Prior works in 3D object detection have modeled these relationships using multi-layer perceptrons~\cite{qi2017pointnet,qi2017pointnet++,qi2018frustum,shi2020pv}, farthest point sampling layers~\cite{qi2017pointnet++,shi2019pointrcnn,shi2020pv}, max pooling layers~\cite{lan2019modeling,qi2017pointnet++,shi2019pointrcnn,shi2020pv}, and graph convolutional networks~\cite{wang2019dynamic}. However, a key challenge is to model these mutual relationships along  with fusing different representations and multi-scale features.

We  present \textsc{M3DETR}, a novel two-stage architecture for 3D object detection task. Given raw 3D point cloud data, our approach can localize static and dynamic obstacles with state-of-the-art accuracy. As illustrated in Figure~\ref{fig:coverpic.p}, the key components of our approach are M3 transformers, which are used to combine different feature representations. Conceptually, each of the M3 transformers are used for aggregating point cloud representations, multi-scale representations and mutual relationships among a subset of points in the point cloud data.
\\

\noindent \textbf{Summary of contributions:}
\begin{itemize}
    \item \textsc{M3DETR} is the first unified architecture for 3D object detection with transformers that accounts for multi-representation, multi-scale, mutual-relation models of point clouds in an end-to-end manner.

    \item \textsc{M3DETR} is robust and insensitive with respect to the hyper-parameters of transformer architectures. 
    We test multiple variants with different transformer blocks designs.
    We demonstrate improved performance of \textsc{M3DETR}  regardless of hyper-parameters. 

    \item Our unified architecture achieves state-of-the-art performance on KITTI 3D Object Detection Dataset~\cite{geiger2012we} and Waymo Open Dataset~\cite{Sun_2020_CVPR}. We outperform the previous state-of-the-art approaches by 2.86\% mAP for car class on the Waymo validation set and 1.48\% mAP for all classes on the Waymo test set.
\end{itemize}

\section{Related Work}
\textbf{Multi-representation modeling.}
Existing techniques for modeling 3D point cloud data include bird-eye-view (BEV), volumetric, and point-wise representations. 
Generally, BEV-based approaches ~\cite{chen2017multi, ku2018joint, liang2018deep, liang2019multi} first project 3D point clouds into 2D BEV space and then adopt the standard 2D object detectors to generate 3D object proposals from projected 2D feature maps. To deal with the irregular format of input point clouds, voxel-based architectures~\cite{zhou2018voxelnet, yan2018second, lang2019pointpillars} use equally spaced 3D voxels to encode the point clouds such that the volumetric representation can be consumed by the region proposal network (RPN) ~\cite{ren2015faster}. Inspired by the PointNet/PointNet++ approach~\cite{qi2017pointnet, qi2017pointnet++}, which is invariant under transformation,~\cite{qi2018frustum, shi2019pointrcnn} extend this method to the task of 3D object detection and directly process the raw point clouds to infer 3D bounding boxes. However, these methods are typically limited due to either information loss or high computation cost. Recently, many approaches~\cite{yang2019std, liu2019point, tang2020searching, shi2020pv, yin2020center} have combined the advantages of speed (of voxel-based representation) and efficiency (of point-based representation) by fusing point-voxel features for 3D object prediction.

\textbf{Multi-scale modeling.} Modeling multi-scale features is an important procedure in deep learning-based computer vision~\cite{lin2017feature, cai2019guided, liu2019l2g, zhang2020feature, liao2020mmnet, zhu18b,qi2017pointnet++, hu2020you, wang2020infofocus} because it is able to enlarge the receptive field and increase resolution. In 3D representation, modeling multi-scale features is also popular and important. PointNet++~\cite{qi2017pointnet++} proposes the set abstraction module to model local features of a cluster of point clouds. To model multi-scale patterns of point clouds, they use 3 different sampling ranges and radii with 3 parallel PointNets and thus fuse the multi-scale. In 3D object detection,~\cite{hu2020you, wang2020infofocus, ye2020hvnet} adopt different detection heads with multi-scale feature maps to handle both large and small object classes. 

\begin{figure*}[t]
    \centering
    \includegraphics[width=\linewidth]{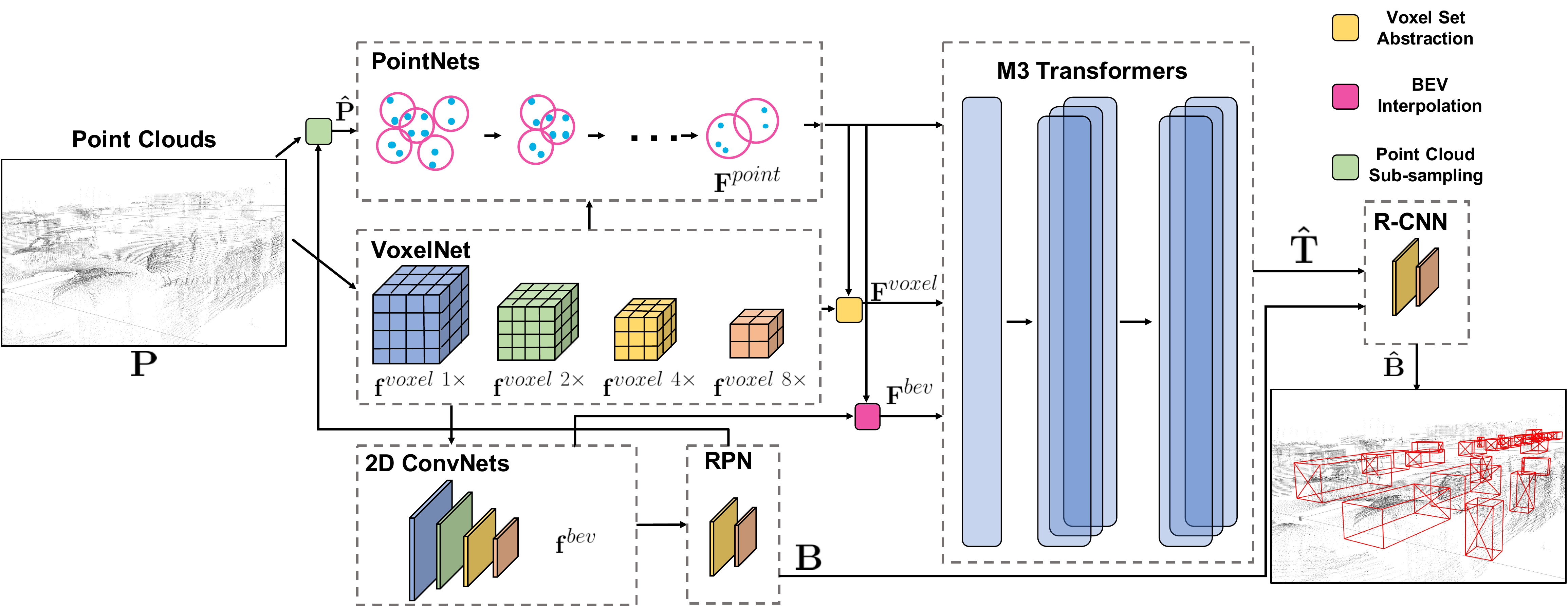}
    \caption{\textbf{An overview of our \textsc{M3DETR} architecture: \textsc{M3DETR} is a transformer based framework for object detection in a coarse-to-fine manner.} It consists of three parts. PointNets, VoxelNet, and 2D ConvNets modules enable individual multi-representation feature learning.  M3 Transformers enable inter-intra multi-representation, multi-scale, multi-location feature attention. With the Region Proposal Network (RPN), the initial box proposals are generated. R-CNN captures and refines region-wise feature representations from M3 transformer output to improve detection performance. } 
    \label{fig:3MDETR_network}
    \vspace{-0.2cm}
\end{figure*}

\textbf{Mutual-relation modeling.} 2D Convolutional Neural Networks \cite{He_2016_CVPR} are commonly used to process mutual relations in 2D images. As point clouds are scattered and lacking structure, passing information from one point to its neighbors is not trivial. PointNets~\cite{qi2017pointnet++} proposes the set abstraction module to model the local context by using the subsampling and grouping layers. After this well-known work, many convolution-like operators~\cite{thomas2019kpconv,lan2019modeling,wang2019dynamic} on point clouds have been proposed to model the local context and the mutual relation between points. Recently, transformers~\cite{zhao2020point,engel2020point,guo2020pct,pan20203d} have been introduced in PointNets to model mutual relation. However, those previous works mainly focus on the local and global contexts of point clouds by applying mutual-relation transformers on points. Instead, our approach not only models the mutual relation between points, but it also models multi-scale and multi-representation features of point clouds.

\textbf{Transformers in computer vision.}
Inspired by their success in machine translation ~\cite{vaswani2017attention}, transformer-based architectures have recently become effective and popular in a variety of computer vision tasks. Particularly, the design of self-attention and cross-attention mechanisms in transformers has been successful in modeling dependencies and learning richer information.~\cite{dosovitskiy2020image} leverages the direct application of transformers on the image recognition task without using convolution.~\cite{carion2020end, zhu2020deformable} apply transformers to eliminate the need for many handcrafted components in conventional object detection and achieve impressive detection results. ~\cite{yang2020learning, zeng2020learning} explore transformers on image and video synthesis tasks.~\cite{zhao2020point} investigates the self-attention networks on 3D point cloud segmentation task. 

Recently, there have been several joint representation fusion research attempts. PointPainting~\cite{vora2020pointpainting} proposes a novel method that accepts both images and point clouds as inputs to do 3D object detection by appending 2D semantic segmentation labels to LiDAR points. In the visual question answering task,~\cite{hu2020iterative} jointly fuses and reasons over three different modality representations. ~\cite{mittal2020m3er} combines multimodal information to solve robust emotion recognition problems. Building on a multi-representation and multi-scale transformer, our proposed model addresses the voxel-wise, point-wise, and BEV-wise feature representation gap and enables effective cross-representation interactions with different levels of semantic features. Coupled with a point-wise mutual-relation transformer, our framework learns to capture deeper local-global structures and richer geometric relationships among point clouds.

\section{Our Approach}
\textsc{M3DETR} takes point cloud data as input and generates 3D boxes for different object categories with state-of-the-art accuracy as shown in Figure~\ref{fig:3MDETR_network}. Our goal is to perform multi-representation, multi-scale, and mutual-relation fusion with transformers over a joint embedding space. Our method consists of three main steps:

\begin{itemize}
    \item Generate feature embeddings for different point cloud representations using VoxelNet, PointNet, and 2D ConvNet.
    
    \item Fuse these embeddings using \textit{M3 transformer} that leverages multi-representation and multi-scale feature embedding and models mutual relationships between points.
    
    \item Perform 3D detection using detection heads network, including RPN and R-CNN stages.
\end{itemize}


\subsection{Multi-Representation Feature Embeddings}
\label{backbone}
Our network processes the raw input point clouds $\bm{P} = \{\bm{p}_1\text{,} \bm{p}_2\text{, ...,} \bm{p}_n\}$ and encodes them into three different embedding spaces, namely, voxel-, point-, and BEV-based feature representations. We discuss the embedding process for each representation in detail.

\noindent\textbf{Voxels:}
Voxel-wise feature extraction is divided into two steps: (1) the voxelization layer, which is used by VoxelNet~\cite{zhou2018voxelnet}, takes the raw input point clouds $\bm{P}$ and converts them into equally spaced 3D voxels $\bm{v} \in \mathbb{R}^{L \times W \times H}$; (2) voxel-wise features $\bm{f}^{voxel}$ at different scales are extracted with 3D sparse convolutions, which is visualized in Figure~\ref{fig:3MDETR_network}. Unlike ~\cite{zhou2018voxelnet, yan2018second}, all four different scales of obtained 3D voxel CNN features, $\bm{f}^{voxel} = \{\bm{f}^{voxel\ 1\times},\  \bm{f}^{voxel\ 2\times},\  \bm{f}^{voxel\ 4\times},\  \bm{f}^{voxel\ 8\times} \}$ are passed to the transformer stage as shown in Figure~\ref{fig:3MDETR_network}. More details will be introduced in the supplementary materials.

\noindent\textbf{Bird's-eye-view:}
2D ConvNets takes the $8 \times$ downsampled voxel-based feature map $\bm{f}^{voxel\ 8\times}$ from the 3D Voxel CNN branch and generates a corresponding BEV-representation embedding. To directly apply 2D convolution, we start by combining the z-dimension and the channel dimension of the input voxel features into a single dimension. The structure of 2D ConvNets includes two encoder-decoder blocks, where each block consists of 2D down-sampling convolutional layers to produce top-down features, as well as de-convolutional layers to upsample to the input feature size. Specifically, both encoder and decoder paths are composed of a number of 2D convolutional layers with the kernel size of $3 \times 3$ followed by a BatchNorm layer and a ReLU layer. The output BEV-based feature from 2D ConvNets, denoted as $\bm{f}^{bev} \in \mathbb{R}^{L_{bev}\times W_{bev}\times C_{bev}}$, was further converted into keypoint feature $\bm{F}^{bev}\in \mathbb{R}^{n\times c_{bev}}$ through bi-linear interpolation. 

\noindent\textbf{Points:}
Typically, there are more than 10K raw points inside an entire point cloud scene. In order to cover the entire point set effectively without large memory consumption, we apply Furthest-Point-Sampling (FPS) algorithm to sample $n$ keypoints, denoted as $\bm{\hat{P}} \subset \bm{P}$. Adopted from PointNet++~\cite{qi2017pointnet++} and PV-RCNN~\cite{shi2020pv}, Set Abstraction and Voxel Set Abstraction (VSA) module take raw point coordinates $\bm{P}$ and the 3D voxel-based features $\bm{f}^{voxel}$, respectively, to generate keypoint features for $\bm{\hat{P}}$. In particular, the corresponding keypoint features from raw points $\bm{P}$ is $\bm{F}^{point} \in \mathbb{R}^{n\times c_{point}}$, and the corresponding keypoint features from voxels $\bm{f}^{voxel}$ are $\{\bm{F}^{voxel\ 1\times}$, $\bm{F}^{voxel\ 2\times}$, $\bm{F}^{voxel 4\times}$,  $\bm{F}^{voxel\ 8\times} \}$, where $\bm{F}^{i}\in \mathbb{R}^{n\times c_{i}}$.

\subsection{Multi-representation, Multi-scale, and Mutual-relation Transformer}
\label{m3transformer}
Once the three feature embedding sequences, $\bm{F}^{voxel}$, $\bm{F}^{point}$ and $\bm{F}^{bev}$ are generated, they are able to dynamically and intelligently attend to each other, generating final cross-representations, cross-scales, and cross-points descriptive feature representations as shown in Figure \ref{fig:3MDETR_network}. In the remainder of this section, we will briefly review transformers basics followed by discussing the multi-representation, multi-scale, and mutual-relation transformer layers. 

\noindent\textbf{Transformer basics.}
A transformer is a stacked encoder-decoder architecture relying on a self-attention mechanism to compute representations of its input and output~\cite{vaswani2017attention}. Consider two input matrices $\bm{X}_l \in \mathbb{R}^{l\times d_{in}}$ and $\bm{X}_s \in \mathbb{R}^{s \times d_{in}}$, where $l$ and $s$ are the lengths of the input sequence of dimension $d_{in}$. The attention layer output is defined as: \vspace{-3.5 mm}

$$\textrm{Attention}(\bm{Q}, \bm{V}, \bm{K}) = \textrm{Softmax}(\frac{\bm{Q}\bm{K}^{T}}{\sqrt{d_{in}}})\bm{V} \in \mathbb{R}^{l \times d_{out}},$$ 

$\textrm{where}\  \bm{Q} = \bm{X}_l \bm{W}_q,\  \bm{K} = \bm{X}_l \bm{W}_k\  and\  \bm{V} = \bm{X}_s \bm{W}_v.$

The matrices $\bm{Q}$, $\bm{K}$ and $\bm{V}$ represent query, key and value respectively, obtained through projection matrices $\bm{W}_q, \bm{W}_k\in \mathbb{R}^{d_{in}\times d_{h}}$ and $\bm{W}_v\in \mathbb{R}^{d_{in}\times d_{out}}$, where $d_{h}$ is the hidden dimension, and $d_{out}$ is the output dimension. When two input matrices $\bm{X}_l$ and $\bm{X}_s$ represent the same feature maps, the attention module is usually referred to as self-attention. Each transformer layer consists of one Multi-head Self-attention (MHSA) module and a few linear layers, as well as normalization and activation layers. More details will be introduced in the supplementary materials.

\begin{figure}[t]
    \centering
    \includegraphics[width=0.85\linewidth]{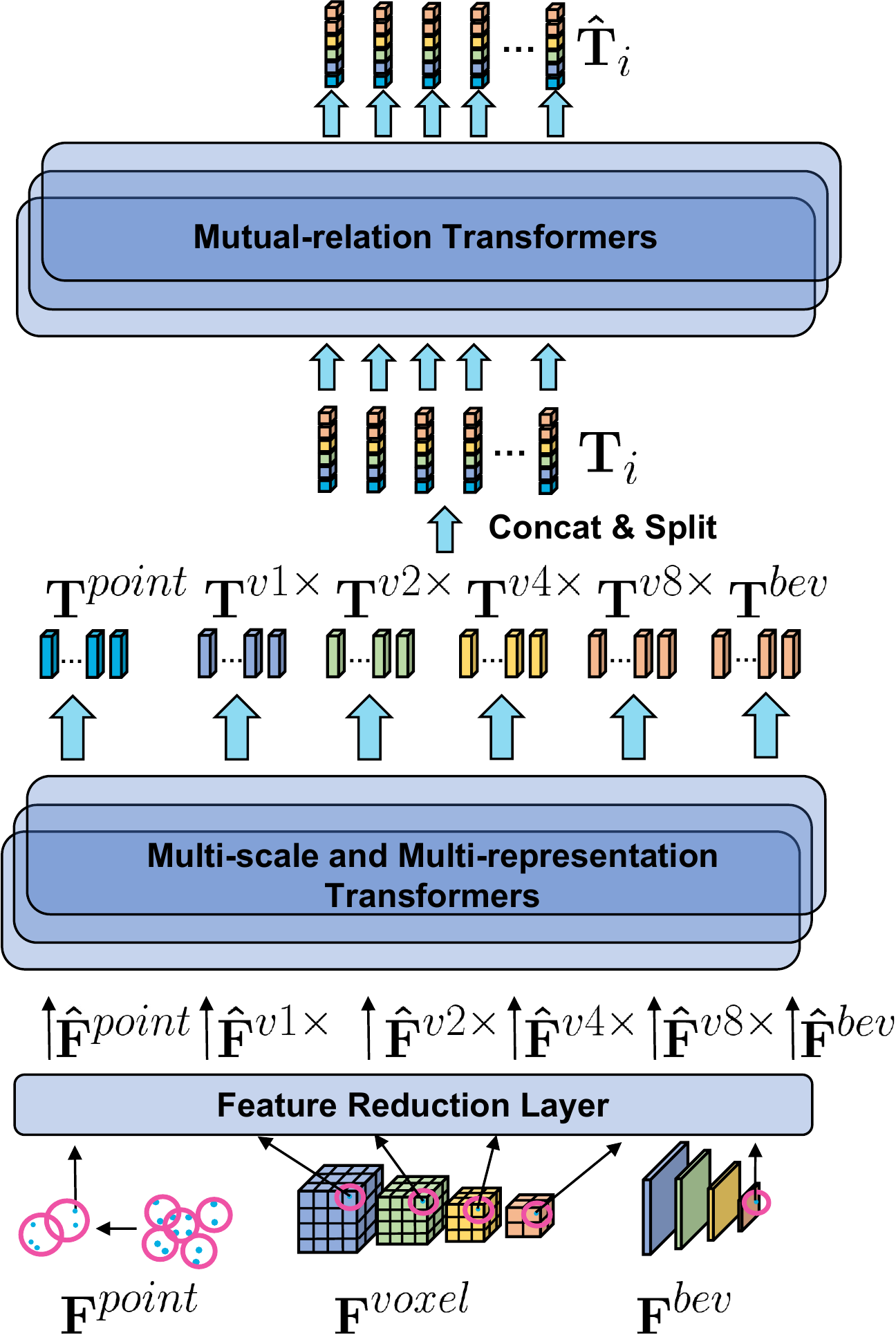}
    \caption{\textbf{M3 Transformers} consist of two parts: a multi-representation and multi-scale transformer, and a mutual-relation transformer. Multi-representation and multi-scale transformer takes the different feature embedding and generates enriched cross-representations and cross-scales embedding. On top of it, the mutual-relation transformer further models point-wise feature relationship to extract the refined features.}
    \label{fig:3MDETR_transformer}
    \vspace{-0.3cm}
\end{figure}




Now, we present the proposed transformers used to capture the inter- and intra- interactions among input features. Specifically, we propose two stacked transformer encoder layers named M3 Transformers, as shown in Figure \ref{fig:3MDETR_transformer}: (1) the multi-representation and multi-scale transformer, and (2) the mutual-relation transformer.

\noindent\textbf{Multi-representation and multi-scale transformer layer.}
First, we focus on the intra-point representation fusion. For each individual point, the input sequence to this transformer layer is its several different corresponding point cloud features, including various scales and distinctive representations. The input sequence to the transformer is $\bm{F} = [\bm{F}^{voxel\ 1\times}$, $ \bm{F}^{voxel\ 2\times}$, $ \bm{F}^{voxel\ 4\times}$, 
$\bm{F}^{voxel\ 8\times}$, $\bm{F}^{point}$, $\bm{F}^{bev}]$, where each $\bm{F}^{i}\in \mathbb{R}^{n\times c_i}$. After explicitly modeling all element-wise interactions among those features for each point separately, the output from the block is the updated feature vectors after aggregating the information from all the input sequence $\bm{F}$.

The multi-representation and multi-scale transformer layer takes 6 different inputs: $\bm{F}^{point}$, $\bm{F}^{voxel 1\times}$, $\bm{F}^{voxel 2\times}$, $\bm{F}^{voxel 4\times}$, $\bm{F}^{voxel 8\times}$,  $\bm{F}^{bev}$. As different inputs may have different feature dimensions, we use the single-layer perceptron to apply feature reduction on the input features to align the feature dimensions of each feature embeddings. The outputs of the feature reduction layer are $\bm{\hat{F}} = [\bm{\hat{F}}^{voxel\ 1\times}$, $ \bm{\hat{F}}^{voxel\ 2\times}$, $ \bm{\hat{F}}^{voxel\ 4\times}$, $\bm{\hat{F}}^{voxel\ 8\times}$, $\bm{\hat{F}}^{point}$, $\bm{\hat{F}}^{bev}]$, where the output dimension of each feature is equivalent to $\hat{c}$.

After the feature reduction layer, the multi-representation and multi-scale transformer layer takes $\bm{\hat{F}}$ as inputs and generates self-attention features $\bm{T}^{voxel\ 1\times}$, $\bm{T}^{voxel\ 2\times}$,$\bm{T}^{voxel\ 4\times}$,$\bm{T}^{voxel\ 8\times}$, $\bm{T}^{point}$, $\bm{T}^{bev}$, which corresponds to $\bm{\hat{F}}^{voxel\ 1\times}$, $\bm{\hat{F}}^{voxel\ 2\times}$,$\bm{\hat{F}}^{voxel\ 4\times}$,$\bm{\hat{F}}^{voxel\ 8\times}$, $\bm{\hat{F}}^{point}$, $\bm{\hat{F}}^{bev}$. We visualize the multi-representation and multi-scale transformer in Figure \ref{fig:3MDETR_transformer}.

\noindent\textbf{Mutual-relation transformer layer.}
Inspired by~\cite{qi2017pointnet, qi2017pointnet++, pan20203d}, inter-points feature fusion within a spatial neighboring space is leveraged in the mutual-relation layer of the transformer. Our goal is to attend to and aggregate the neighboring information in an attention manner for each point with an enriched feature.

From the first transformer block output, we obtain aggregated features $\bm{T}$ of different scales and representations after the first transformer block. We concatenate channels along the point dimension and rearrange those points into a sequence as the input of the mutual-relation transformer. Let the learned concatenated feature $\bm{T} = 
\rm{concat}$ $\{\bm{T}^{voxel\ 1\times}$, $ \bm{T}^{voxel\ 2\times}$, $ \bm{F}^{voxel\ 4\times}$, $\bm{T}^{voxel\ 8\times}$, $\bm{T}^{point}$, $\bm{T}^{bev}\} \in\mathbb{R}^{n\times c_T}$, where $c_{T}$ is the sum of all channel size in $\bm{T}$.  Note that, the first dimension of $\bm{T}$ is the number of keypoints $n$. To model the mutual-relation between keypoints, we need to split $\bm{T}$ into $n$ point-wise feature $\bm{T} = \{ \bm{T}_i\}$, where $\bm{T}_i \in \mathbb{R}^{c_{T}}$. The concatenation and split module is showed in Figure \ref{fig:3MDETR_transformer}.

The mutual-relation transformer takes point-wise features of $n$ keypoints as inputs and uses the multi-head self-attention head to model the mutual relationship between keypoints. The outputs of the mutual-relation transformer are $\bm{\hat{T}}$.

\noindent\textbf{Comparison with previous point-based transformers} 
Prior work has explored the transformer application on the task of point cloud processing. First, \cite{guo2020pct, engel2020point, zhao2020point} leverage the inherent permutation invariance of transformers to capture local context within the point cloud on the shape classification and segmentation tasks, while our \textsc{M3DETR} mainly investigates the strong attention ability of transformers between input embeddings for the 3D object detection task. Pointformer \cite{pan20203d} is the approach most related to our method because we both address the 3D object detection task by capturing the dependencies among points' features. However, Pointformer adopts a single PointNet branch to extract points feature, while \textsc{M3DETR} considers all three different representations and also applies the transformer to learn aggregated representation-based features.

\subsection{Detection Heads Network}
\label{dethead}
After we obtain the enriched embedding $\bm{\hat{T}}$ from the M3 transformer, the detection network is composed of two stages that predict 3D bounding box class, localization, and orientation in a coarse-to-fine manner, including RPN and R-CNN. Please refer to PV-RCNN~\cite{shi2020pv} for more details.

\noindent\textbf{RPN:} Region Proposal Networks take the deep semantic features $\bm{f}^{bev}$ produced by the 2D ConvNets as inputs and generate high-quality 3D object proposals $\bm{B}$, as shown in Figure \ref{fig:3MDETR_network}. A 3D object box $\bm{B}_i$ is parameterized as $(x, y, z, l, h, w, \theta) $, where $(x, y, z)$ is the center of the box, $(l , h, w)$ is the dimension of the box, and $\theta$ is the orientation in bird's-eye-view. Similar to the conventional RPN in 2D object detection~\cite{ren2015faster} each position on the deep feature map is placed by predefined bounding box anchors denoted as ($x^a$, $y^a$, $z^a$, $l^a$, $h^a$, $w^a$, $\theta^a$). Then initial proposals are generated by predicting the relative offset of an object's 3D ground truth bounding box ($x^{gt}$, $y^{gt}$, $z^{gt}$, $l^{gt}$, $h^{gt}$, $w^{gt}$, $\theta^{gt}$).


\noindent\textbf{R-CNN:}
R-CNN serves as the second stage and it takes the initial region proposals $\bm{B}$ from RPN as input to conduct further proposal refinement. For each input proposal box, the RoI-grid pooling module~\cite{shi2020pv} is adopted to extract the corresponding proposal-specific grid points' features from the transformer-based embeddings $\bm{\hat{T}}$. 
Compared with previous works, \textsc{M3DETR} leverages the richer embedding information from the learned transformers for the fine-grained proposal refinement. As the main component to extract refined features, the RoI-grid pooling module uniformly samples $ N \times N \times N $ grid points per 3D proposal. For each grid point, the output feature is generated by applying a PointNet-block \cite{qi2017pointnet} on a small number of surrounding keypoints, $M$, within its spatial surrounding region with a radius of $r$. Specifically, keypoints are the subset of input points that are sampled using the Furthest-point-sampling algorithm to cover the entire point set. 

\begin{table*}[ht]
    \centering
\begingroup

\setlength{\tabcolsep}{2.4pt} 
\renewcommand{\arraystretch}{1.2} 
\resizebox{0.98\linewidth}{!}{%

\begin{tabular}{l | c c c c | c c c c | c c c c | c c c c}
\toprule
 & \multicolumn{4}{c|}{Vehicle} & \multicolumn{4}{c|}{Pedestrian} &  \multicolumn{4}{c|}{Cyclist} & \multicolumn{4}{c}{All} \\
Method & L1 mAP & L1 mAPH
& L2 mAP & L2 mAPH & L1 mAP & L1 mAPH
& L2 mAP & L2 mAPH & L1 mAP & L1 mAPH
& L2 mAP & L2 mAPH & L1 mAP & L1 mAPH
& L2 mAP & L2 mAPH\\
\hline
PV-RCNN* \cite{shi2020pv} & 76.89 & 76.30 & 67.99 & 67.46 &
65.43 & 55.85 & 59.56 & 50.74 &
66.38 & 64.68 &  63.42 & 61.81 & 
69.57 & 65.61 & 63.65  & 60.00 \\
\hline


\textsc{M3DeTR} & 77.66 & 77.09 & 70.54 & 69.98 &
68.20 & 58.50 & 60.64 & 52.03 & 
67.28 & 65.69 &  65.31 & 63.75 & 
71.05 & 67.09 & 65.50  & 61.92 \\

Improvement & \textbf{+0.77} & \textbf{+0.79} & \textbf{+2.55} & \textbf{+2.52} & 
\textbf{+2.77} & \textbf{+2.65} & \textbf{+1.08} & \textbf{+1.29} & 
\textbf{+0.90} & \textbf{+1.01} & \textbf{+1.89} & \textbf{+1.94} &  
\textbf{+1.48} & \textbf{+1.48} & \textbf{+1.85} & \textbf{+1.92} \\
\bottomrule
\end{tabular}
}
\vspace*{1mm}
\caption{\textsc{M3DETR} outperforms in the Vehicle, Pedestrian and Cyclist classes for both LEVEL\_1 and LEVEL\_2 difficulty levels on Waymo Open Dataset test set. Note that PV-RCNN* is our reproduced results with single point cloud frame input.}
\vspace*{-1mm}
\label{tab:waymo_test_all_classes}
\endgroup

\end{table*}
\begin{table*}[ht]
    \centering
\begingroup

\setlength{\tabcolsep}{2.4pt} 
\renewcommand{\arraystretch}{1.2} 
\resizebox{0.98\linewidth}{!}{%

\begin{tabular}{l | c c c c | c c c c | c c c c | c c c c}
\toprule
 & \multicolumn{4}{c|}{3D mAP LEVEL\_1} & \multicolumn{4}{c|}{3D mAPH LEVEL\_1} & \multicolumn{4}{c|}{3D mAP LEVEL\_2} & \multicolumn{4}{c}{3D mAPH LEVEL\_2}\\
Method & Overall
& 0-30m & 30-50m  & 50m-Inf & Overall
& 0-30m & 30-50m  & 50m-Inf & Overall & 0-30m & 30-50m  & 50m-Inf & Overall
& 0-30m & 30-50m  & 50m-Inf\\
\hline
RCD~\cite{bewley2020range} &  69.59 & 87.2 & 67.8 & 46.1 & - & - & - & - & - & - & - & -  & - & - & - & - \\
StarNet~\cite{ngiam2019starnet} & 61.50 & 82.20 & 56.60 & 32.20 & 61.00 & 81.70 & 56.00 & 31.80 & 54.9 & 81.3 & 49.5 & 23.0 & 54.50 & 80.80 & 49.00 & 22.70 \\
PointPillars~\cite{lang2019pointpillars} & 68.62 & 87.20 & 65.50 & 40.92 & 68.08 & 86.71 & 64.87 & 40.19  & 65.21 & 87.93 & 63.80 & 38.20 & 64.29 & 87.30 & 62.36 & 36.87  \\
Det3D~\cite{zhu2019class} & 73.29 & 90.31 & 70.54 & 49.10 & 72.27 & 89.65 & 68.96 & 47.45  & 65.21 & 87.93 & 63.80 & 38.20 & 64.29 & 87.30 & 62.36 & 36.87 \\
RangeDet~\cite{fan2021rangedet} & 75.83 & 88.41 & 73.83 & 55.31 & 75.38 & 87.95 & 73.38 & \underline{54.84} & 67.12 & 87.53 & 67.99 & \underline{44.40} & 66.73 & 87.08 & 67.58 & \underline{44.01} \\
PV-RCNN* \cite{shi2020pv} & \underline{76.89} & \underline{92.27} & \underline{75.51} & \underline{55.35} & \underline{76.30} & \underline{91.82} & \underline{74.79} & 54.27  & \underline{67.99} & \underline{89.18} & \underline{69.39} & 42.80 & \underline{67.46} & \underline{88.75} & \underline{68.70} & 41.95 \\


\hline
\textsc{M3DeTR} & \textbf{77.66} & \textbf{92.54} & \textbf{76.27} & \textbf{57.12} &
\textbf{77.09} & \textbf{92.09} & \textbf{75.61} & \textbf{56.02} & 
\textbf{70.54} & \textbf{89.43} & \textbf{70.19} & \textbf{45.57} &
\textbf{69.98} & \textbf{89.01} & \textbf{69.54} & \textbf{44.62}  \\
Improvement & \textbf{+0.77} & \textbf{+0.27} & \textbf{+0.76} & \textbf{+1.77} &
\textbf{+0.79} & \textbf{+0.27} & \textbf{+0.82} & \textbf{+1.18} & 
\textbf{+2.55} & \textbf{+0.25} & \textbf{+0.80} & \textbf{+1.17} &
\textbf{+2.52} & \textbf{+0.26} & \textbf{+0.84} & \textbf{+0.61}  \\

\bottomrule
\end{tabular}
}
\vspace*{1mm}
\caption{\textsc{M3DETR} outperforms in the Vehicle class with different size range on Waymo Open Dataset test set. Note that PV-RCNN* is our reproduced results with single point cloud frame input. We underscore the second best method in each column for comparison.}
\label{tab:waymo_test}
\vspace*{-6mm}
\endgroup

\end{table*}
\begin{table*}[ht]
    \centering
\begingroup

\setlength{\tabcolsep}{2.4pt} 
\renewcommand{\arraystretch}{1.2} 
\resizebox{0.98\linewidth}{!}{%

\begin{tabular}{l | c c c c | c c c c | c c c c | c c c c}
\toprule
 & \multicolumn{4}{c|}{3D mAP LEVEL\_1} & \multicolumn{4}{c|}{3D mAPH LEVEL\_1} & \multicolumn{4}{c|}{3D mAP LEVEL\_2} & \multicolumn{4}{c}{3D mAPH LEVEL\_2}\\
Method & Overall
& 0-30m & 30-50m  & 50m-Inf & Overall
& 0-30m & 30-50m  & 50m-Inf 
& Overall
& 0-30m & 30-50m  & 50m-Inf & Overall
& 0-30m & 30-50m  & 50m-Inf\\
\hline
LaserNet~\cite{meyer2019lasernet} &  52.11 & 70.90 & 52.90 & 29.60 & 50.05 & 68.70 & 51.40 & 28.60 & - & - & - & - & - & - & - & - \\
PointPillars~\cite{lang2019pointpillars} & 56.62 & 81.00 & 51.80 & 27.90 & - & - & - & - & - & - & - & - & - & - & - & - \\
RCD~\cite{bewley2020range} & 69.59 & 87.20 & 67.80 & 46.10 & 69.16 & 86.80 & 67.40 & \underline{45.50} & - & - & - & - & - & - & - & - \\
RangeDet \cite{fan2021rangedet} & \underline{72.85} & 87.96 & 69.03 & \underline{48.88}  & - & - & - & - & - & - & - & - & - & - & - & -\\
PV-RCNN \cite{shi2020pv} & 70.30  & \underline{91.90}  & \underline{69.20} & 42.20 & \underline{69.69}  & \underline{91.34} & \underline{68.53} & 41.31 & \underline{65.36}  & \underline{91.58} & \underline{65.13} & \underline{36.46} & \underline{64.79}  & \underline{91.00} & \underline{64.49} & \underline{35.70} \\


\hline
\textsc{M3DeTR} &  \textbf{75.71} & \textbf{92.69}  & \textbf{73.65}  & \textbf{52.96}  & 
\textbf{75.08} & \textbf{92.22} & \textbf{72.94} & \textbf{51.80} & 
\textbf{66.58} &\textbf{ 91.92}  & \textbf{65.73}  & \textbf{40.44}   & 
\textbf{66.02} & \textbf{91.45} & \textbf{65.10} & \textbf{39.52} \\
Improvement & \textbf{+2.86} & \textbf{+0.79}  & \textbf{+4.45}  & \textbf{+3.98}  & 
\textbf{+5.39} & \textbf{+0.88} & \textbf{+4.41} & \textbf{+6.3} & 
\textbf{+1.22} & \textbf{+0.34} & \textbf{+0.6} & \textbf{+3.94} & 
\textbf{+1.23} & \textbf{+0.45} & \textbf{+0.61} & \textbf{+3.82}  \\
\bottomrule
\end{tabular}
}
\vspace*{1mm}
\caption{\textsc{M3DETR} outperforms in the Vehicle class with IoU threshold of 0.7 on the full 202 Waymo Validation Set, especially on the far range (50m to Inf). We underscore the second best method in each column for comparison. }
\label{tab:waymo_val}
\vspace*{-4mm}
\endgroup

\end{table*}


Finally, the refined representations of each proposal are first passed to two fully connected layers and then generate the box 3D Intersection-over-Union (IoU) guided confidence scoring of class prediction and location refinement of regression targets, $\bm{\hat{B}}$. Compared with the traditional classification-guided box scoring, 3D IoU guided confidence scoring considers the IoU between the proposal box and its corresponding ground truth box. Empirically, \cite{shi2020points,shi2020pv} show that it achieves better results compared with the traditional classification confidence based techniques. \vspace{-2 mm}

\subsection{Loss Functions} 
\label{loss}
In this section, we define the loss function. The bounding box regression target for both RPN and R-CNN stages is calculated as the relative offsets between the anchors and the ground truth as:
$\triangle x = \frac{x^{gt} - x^a}{d^a},  \triangle y = \frac{y^{gt} - y^a}{d^a}$,  
$\triangle z = \frac{z^{gt} - z^a}{h^a}$
$\triangle h = \log (\frac{h^{gt}}{{h^{a}}})$,  
$\triangle w = \log (\frac{w^{gt}}{{w^{a}}})$,
$\triangle \theta = \theta^{gt} - \theta^{a}$,
 where $ d^a $ = $\sqrt{(w^a)^2 + (l^a)^2} $. Similar to~\cite{shi2020pv, shi2020points}, the focal loss is applied \cite{lin2017focal} for the classification loss, $\mathcal{L}_{cls}$. Smooth L1 loss \cite{girshick2014rich} is adopted for the box localization regression target's losses, $\mathcal{L}_{reg}$ and $\mathcal{L}_{ref}$. In addition, 3D IoU loss \cite{shi2020pv, shi2020points} is used for $\mathcal{L}_{iou}$. 

Similar to PV-RCNN \cite{shi2020pv}, we formally define a multi-task loss for both the RPN and R-CNN stages, 
\vspace{1pt}
\begin{equation}
\begin{aligned}
 & \mathcal{L}_{cls} = -\alpha_{a}(1 - p^{a})^\gamma \log p^{a}, \\
 & \mathcal{L}_{reg} = \sum_{b \in (x, y, z, w, l, h, \theta)} \mathcal{L}_s (\triangle b), \\
 & \mathcal{L} = \mathcal{L}_{cls} + \beta_{reg} \mathcal{L}_{reg} + \beta_{iou} \mathcal{L}_{iou} + \beta_{ref} \mathcal{L}_{ref},
\end{aligned}
\end{equation}

where $p^{a}$ is the model's estimated class probability for an anchor box, and $\beta_{reg}$, $\beta_{iou} $ and $\beta_{ref}$ are chosen to balance the weights between classification loss, IoU loss and regression loss for RPN stage and R-CNN stage. We adopt the default $\alpha = 0.25$, and $\gamma = 2.0$ from the parameters of focal loss~\cite{lin2017focal}.

\section{Experiments}
In this section, we evaluate \textsc{M3DETR} both qualitatively and quantitatively on the Waymo Open Dataset~\cite{Sun_2020_CVPR} and the KITTI Dataset~\cite{geiger2012we} in the task of LiDAR-based 3D object detection. Our main results include achieving a state-of-the-art accuracy on these datasets and robustness to hyper-parameter tuning.


\subsection{Datasets and Evaluation Metric}
\label{dataset}
\noindent \textbf{Waymo Open Dataset:}
The Waymo Open Dataset~\cite{Sun_2020_CVPR} is a large-scale autonomous driving dataset containing $1000$ scenes of $20$s duration each, with $798$ scenes for training and $202$ scenes for validation. Each scene is sampled at a frequency of $10$Hz. Overall, the dataset includes $12$M labeled objects and thus we only use one fifth of the training scenes for the following experiment. We consider LiDAR data as the input to our approach. The evaluation protocol on the Waymo dataset consists of the mean average precision (mAP) and mean average precision weighted by heading (mAPH). For each object category, the detection outcomes are evaluated based on two difficulty levels: LEVEL\_1 denotes the annotated bounding box with more than 5 points and LEVEL\_2 represents the annotated bounding box with more than 1 point.

\textbf{KITTI dataset}
The KITTI 3D object detection benchmark~\cite{geiger2012we} is another popular dataset for autonomous driving. It contains $7,481$ training and $7,518$ testing LiDAR scans. We follow the standard split on the training ($3,712$ samples) and validation sets ($3,769$ samples). For each object category, the detection outcomes are evaluated based on three difficulty levels based on the object size, occlusion state, and truncation level. \vspace{-2.5 mm}

\subsection{Implementation Details}\label{impl}
Closely following the codebase\footnote{https://github.com/open-mmlab/OpenPCDet.}, we use PyTorch to implement our M3 transformer modules and integrate them into the PV-RCNN network~\cite{shi2020pv}. More details about backbone and detection heads network will be introduced in the supplementary materials.




\noindent\textbf{M3 Transformer}
We project the embeddings obtained from the backbone network with different scales and representations to 256 channels, as the input of the multi-representation and multi-scale transformer requires, and project the output features back to their original dimensions before passing into the mutual-relation transformer. Due to the GPU memory constraint, we experiment with two types of MHSA module designs: 2 encoder layers with 4 attention heads and 1 encoder layer with 8 attention heads. 



\begin{figure*}[t]
    \centering
    \includegraphics[width=1\linewidth]{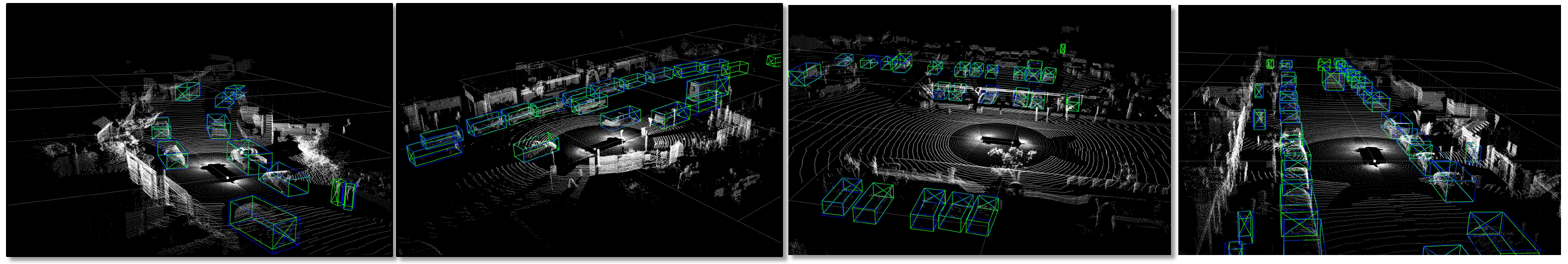}
    \caption{We highlight the 3D detection results of \textbf{M3DETR} on the Waymo Open Dataset. The 3D ground truth bounding boxes are in green, while the detection bounding box are shown in blue.}
    \label{fig:visualization}
    \vspace{-0.4cm}
\end{figure*}

\noindent\textbf{Training Parameters}
Models are trained from scratch on $4$ NVIDIA P6000 GPUs. We use the Adam optimizer with a fixed weight decay of $0.01$ and use a one-cycle scheduler proposed in~\cite{smith2018disciplined}. For the Waymo Open Dataset, we train our models for $45$ epochs with a batch size of $8$ scenes and a learning rate $0.01$, which takes around $50$ hours. For the KITTI dataset, we train our models for $80$ epochs with a batch size of $8$ scenes per and a learning rate $0.01$, which takes around $15$ hours. 

\subsection{Results}
\label{result}
\noindent\textbf{Waymo Open Dataset:}
We first present our object detection results for the vehicle, pedestrian, and cyclist classes on the test set of Waymo Open Dataset in Table~\ref{tab:waymo_test_all_classes} compared with PV-RCNN~\cite{shi2020pv}. We evaluate our method at both LEVEL\_1 and LEVEL\_2 difficulty levels. Note that we reproduce the baseline PV-RCNN with a single frame input since they recently adopt two-frames input on test set. As we can see, PV-RCNN achieves 69.57\% and 63.65\% on average in LEVEL\_1 mAP and LEVEL\_2 mAP, respectively, while \textsc{M3DETR} improves them by 1.48\% and 1.85\%, respectively. Without bells and whistles, our approach works better than PV-RCNN~\cite{shi2020pv}. Furthermore, we compare our framework on the vehicle class for different distances with state-of-the-art methods, including StarNet~\cite{ngiam2019starnet}, PointPillars~\cite{lang2019pointpillars}, RCD~\cite{bewley2020range}, Det3D~\cite{zhu2019class}, RangeDet~\cite{fan2021rangedet} and PV-RCNN~\cite{shi2020pv}. In Table~\ref{tab:waymo_test}, \textsc{M3DETR} outperforms PV-RCNN significantly in both LEVEL\_1 and LEVEL\_2 difficulty levels across all distances, demonstrating the effectiveness of newly proposed framework. Moreover, we visualize the detection results of \textsc{M3DETR} in Figure ~\ref{fig:visualization}.

\begin{figure}[t]
\vspace{-0.5cm}
    \centering
    \centerline{
    \includegraphics[width=1\linewidth]{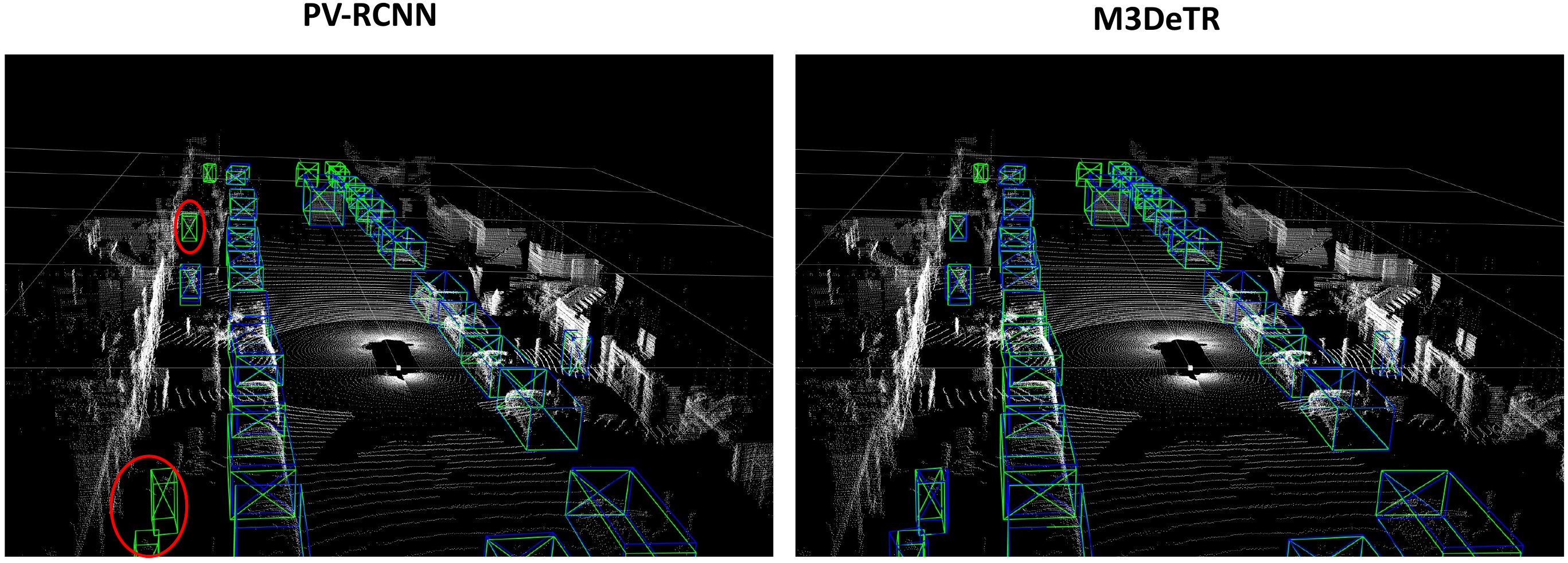}
    }
    \vspace{-1mm}
    \caption{We visualize the 3D detection results for the same input point cloud between PV-RCNN (left) and \textbf{M3DETR} (right) on Waymo Open Dataset. We highlight the false negative boxes from PV-RCNN in red.}
    \label{fig:visualization_comparison}
    \vspace{-0.4cm}
\end{figure}

Compared with the PV-RCNN shown in Figure~\ref{fig:visualization_comparison}, \textsc{M3DETR} successfully captures the inter- and intra- interactions among input features and effectively helps the model generate high-quality box proposals. To the best of our knowledge, \textsc{M3DETR} achieves the state-of-the-art in the Vehicle class in both LEVEL\_1 and LEVEL\_2 difficulty levels among all the published papers with a single frame LiDAR input.

We also evaluate the overall object detection performance with an IoU of 0.7 for Vehicle class on the full Waymo Open Dataset validation set as in Table~\ref{tab:waymo_val}, further proving that our architecture is more efficient for jointly modeling the input features.

\noindent\textbf{KITTI dataset:}
We compare our approach with the state-of-the-art methods on the KITTI test set~\cite{shi2020pv,chen2020object}. We compute the mAP on three difficult types of both car and  cyclist classes in 3D detection metric. Table~\ref{tab:kitti_test_car_cyclist} shows that \textsc{M3DETR} achieves state-of-the-art performance and outperforms the previous work by a large margin especially on the cyclist class. In particular, HotSpotNet~\cite{chen2020object} achieves $82.59\%$ in the ``easy'' categories of 3D detection metric, while \textsc{M3DETR} improves these results by significant $1.24\%$.  \vspace{-2 mm}

\begin{table}[t]
    \centering
    \setlength\tabcolsep{3px}
\begingroup

\setlength{\tabcolsep}{2.4pt} 
\renewcommand{\arraystretch}{1.2} 
\vspace*{-3mm}

\begin{tabular}{l | c c c | c c c}
\toprule
 & \multicolumn{3}{c|}{Car} & \multicolumn{3}{c}{Cyclist}\\
Method & Easy & Mod & Hard & Easy & Mod & Hard\\
\hline
F-PointNet~\cite{qi2018frustum} &  72.27 & 56.12  & 49.01 & 82.19 & 69.79 & 60.59 \\
VoxelNet~\cite{zhou2018voxelnet} &77.47 & 65.11 & 57.73 & 61.22 & 48.36  & 44.37 \\
SECOND~\cite{yan2018second} & 83.34 & 72.55 & 65.82 & 75.83  & 60.82 & 53.67 \\
PointPillars~\cite{lang2019pointpillars} & 82.58 & 74.31 & 68.99 & 77.10 &  58.65  & 51.92\\
PointRCNN~\cite{shi2019pointrcnn} & 86.96 & 75.64 & 70.70 & 74.96 & 58.82 & 52.53  \\
STD~\cite{yang2019std} & 87.95 & 79.71 & 75.09 & 78.69 & 61.59  & 55.30    \\
HotSpotNet~\cite{chen2020object} & 87.60 & 78.31  & 73.34 & \underline{82.59} &  \underline{65.95}  &  \underline{59.00} \\
PVRCNN~\cite{shi2020pv} &  \underline{90.25} & \underline{81.43} & \underline{76.82}  &  78.60  & 63.71 & 57.65 \\
\hline
\textsc{M3DeTR} & \textbf{90.28} & \textbf{81.73}  & \textbf{76.96}  & \textbf{83.83} & \textbf{66.74}  & \textbf{59.03}  \\
Improvement & \textbf{+0.03} & \textbf{+0.3}  & \textbf{+0.14} & \textbf{+1.24} & \textbf{+0.79}  & \textbf{+0.03} \\



\bottomrule
\end{tabular}
\vspace*{-1mm}
\caption{\textsc{M3DETR} outperforms in both Car and Cyclist classes for 3D detection benchmark on KITTI Test Set. We underscore the second best method in each column for comparison.}
\label{tab:kitti_test_car_cyclist}
\vspace*{-1mm}
\endgroup
\vspace{-0.2cm}
\end{table}

\begin{table}[t]
    \centering
    \setlength\tabcolsep{3px}
\begingroup

\resizebox{0.95\columnwidth}{!}{%

\begin{tabular}{c c | c c c | ccc}
\toprule
 & & \multicolumn{3}{c|}{Recall\_11} & \multicolumn{3}{c}{Recall\_40}\\
Rel. Trans. & Rep. and Scal. Trans. & Easy & Mod & Hard  & Easy & Mod & Hard \\
\hline
x & x & 88.66 & 79.07 & 78.49 & 91.17 & 82.61 & 82.06\\

\checkmark & x & 88.82 & 83.23 & 78.64 & 91.37 & 84.40 & 82.34 \\
x & \checkmark & 88.93 & 83.63 & 78.59 & 91.72 & 84.68 & 82.39 \\
\checkmark  & \checkmark  & \textbf{89.28} & \textbf{84.16} & \textbf{79.05} & \textbf{92.29  
} & \textbf{85.41} & \textbf{82.85} \\ 

\bottomrule
\end{tabular}
}
\vspace*{1mm}
\caption{Ablation studies of transformers in Car class with IoU threshold of 0.7 on KITTI Validation dataset. ”Rel. Trans.” and ”Rep. and Scal. Trans.” refer to mutual-relation trans-former of 2 MHSA layer with 4 heads, and multi-representation and multi-scale transformer of 1 MHSA layers with 8 heads, respectively.}
\label{tab:kitti_ablation_transformer}
\endgroup
\vspace{-0.5cm}
\end{table}

\subsection{Ablation Studies}
\label{ablation}
To demonstrate the individual benefits of the multi-representation, multi-scale, and mutual-relation layers of the M3 transformer, we perform ablation experiments and tabulate the results in Table~\ref{tab:kitti_ablation_transformer}. All experiments are conducted on the validation set of KITTI dataset.

With the single multi-representation and multi-scale transformer layer, we can achieve 4.07\% and 1.71\% on the moderate difficulty in car class with 11 and 40 recall positions, respectively compared with the PV-RCNN baseline. On the other side, with the single mutual-relation transformer layer, the performance gain are 4.47\% and 1.99\% compared with the PV-RCNN baseline. Without hyper-parameter tuning, \textsc{M3DETR} benefits from unifying multiple point cloud representations, feature scales, and model mutual-relations simultaneously which results in the best performance.  \vspace{-2 mm}

\subsection{Robustness of M3DETR}
\label{robust}
\begin{figure}[t]
    \centering
    \includegraphics[width=0.95\linewidth]{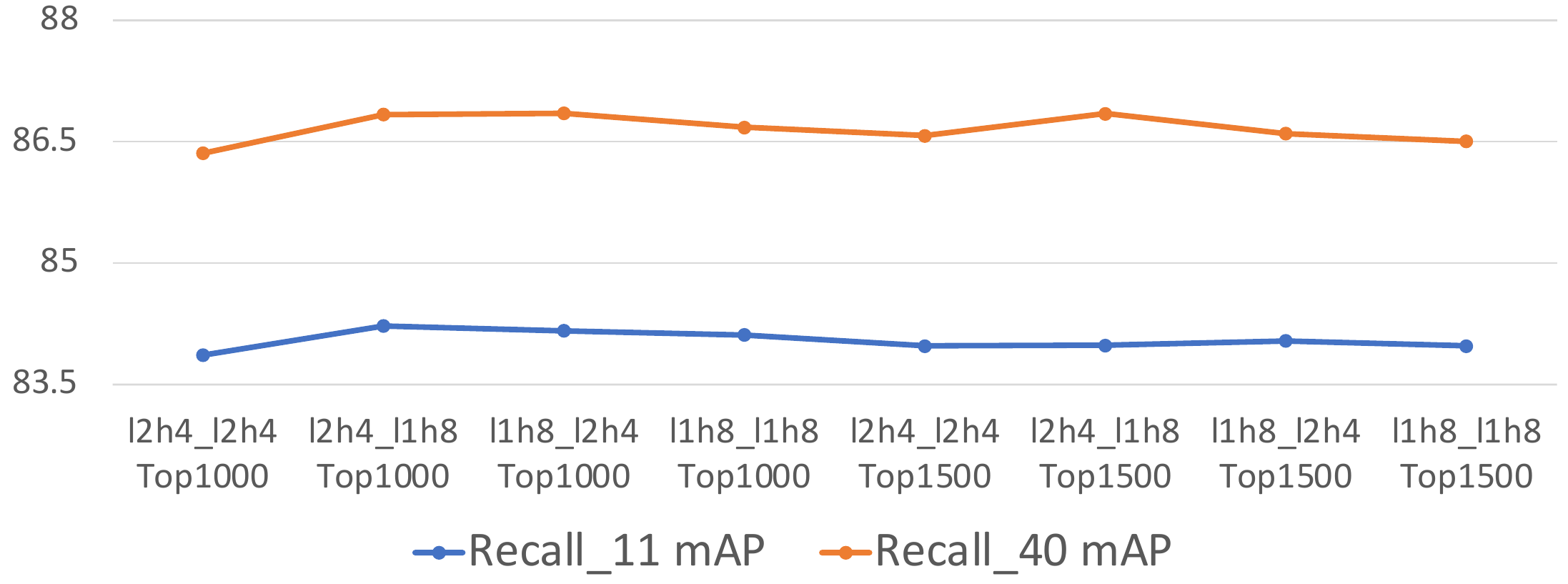}
    \vspace{-1mm}
    \caption{Performance comparison of different \textbf{M3 Transformers} variants in Car class on KITTI validation set. We show the mAP results with IoU threshold of 0.7 for both 11 and 40 recall positions.  Note that "l" and "h" represent layer number and head dimension of M3 Transformers, respectively. "Top" denotes the number of proposals used for keypoint sampling from RPN stage.}
    \label{fig:M3DETR_robustness}
    \vspace{-0.6cm}
\end{figure}

To demonstrate the robustness of \textsc{M3DETR} to hyper-parameter tuning, we perform a series of tests by varying the sampling size, number of detection heads, and number of transformer encoder layers. We present the results of these tests in Figure~\ref{fig:M3DETR_robustness}, where we observe that \textsc{M3DETR} performs consistently well for the ``car'' category with IoU threshold of 0.7 for both 11 and 40 recall positions on the KITTI validation set.

\section{Conclusions}
In this paper, we present \textsc{M3DETR}, a novel transformer-based framework for object detection with LiDAR point clouds. \textsc{M3DETR} is designed to simultaneously model multi-representation, multi-scale, mutual-relation features through the proposed M3 Transformers. Overall, the first transformer integrates features with different scales and representations, and the second transformer aggregates information from all keypoints. Experimental results show that \textsc{M3DETR} outperforms previous work by a large margin on the Waymo Open Dataset and the KITTI dataset. Without bells and whistles, \textsc{M3DETR} is demonstrated to be invariant to the hyper-parameters of transformer. 


\noindent\textbf{Acknowledgement.} This work was supported in part by ARO Grants W911NF1910069, W911NF2110026  and U.S. Army Grant No. W911NF2120076.

\pagebreak


{\small
\bibliographystyle{ieee_fullname}
\bibliography{egbib}
}

\clearpage

\appendix

\section{More Our Approach Details}
We further discuss our approach in the following. 

\subsection{Voxel Representation in Multi-Representation Feature Embeddings}
For the voxel-wise feature extraction from raw point clouds input, there are two steps, voxelization using voxelization layer and feature extraction using 3D sparse convolutions. We denote the size of each discretized voxel as $L \times W \times H \times C$, where $L, W, H$ indicate the length, width, and height of the voxel grid and $C$ represents the channel of the voxel features. We adopt the average of the point-wise features from all the points to represent the whole non-empty voxel feature. After voxelization, the input feature is propagated through a series of $3 \times 3 \times 3$ sparse cubes, including four consecutive blocks of 3D sparse convolution with downsampled sizes of $1 \times$, $2 \times$, $4 \times$, $8 \times$, using convolution operations of stride 2. Specifically, each sparse convolutional block includes a 3D convolution layer followed by a LayerNorm layer and a ReLU layer.





\subsection{Multi-head Self-attention Basics}
Building on the attention mechanism, Multi-head Self-attention (MHSA) with $N$ heads and the input matrix $\bm{X}$ is defined as follows: \vspace{-5mm}

\[\textrm{MHSA}(\bm{Q}, \bm{K}, \bm{V}) = \textrm{Concatenate} [\textrm{head}_1, ..., \textrm{head}_N] \bm{W} ^O, \]\vspace{-3mm}


\noindent where $ \textrm{head}_i = \textrm{Attention}(\bm{QW}_{i}^{q}, \bm{KW}_{i}^{k}, \bm{VW}_{i}^{v})$ and $\bm{Q}, \bm{K}, \bm{V}$ are linear projections of X. The projection matrices $\bm{W}^{q}_i \in \mathbb{R}^{d_h\times d_{q}},\  \bm{W}^{k}_i \in \mathbb{R}^{d_h\times d_{k}},\  \bm{W}^{v}_i \in \mathbb{R}^{d_h\times d_{v}},\  \bm{W}^O \in \mathbb{R}^{d_v\times d_{out}}$ are learnable parameters in the network that correspond to each of the attention heads, where $d_k$, $d_v$ and $d_q$ are the hidden dimensions of each attention head for $\bm{K}$, $\bm{V}$, and $\bm{Q}$, respectively. 

\section{More Implementation Details}
As mentioned in the Section 4.2, we give more details on our implementation for reproduction of our result. Our code will also be released later, including trained models that can match the performance that was included in the paper. 

\subsection{Backbone}
The 3D voxel CNN branch consists of 4 blocks of 3D sparse convolutions with output feature channel dimensions of 16, 32, 64, 64. Those 4 different voxel representations of different scales, as well as point features from point cloud input, are used to refine keypoint features by PointNet~\cite{qi2017pointnet++} through set abstraction and voxel set abstraction~\cite{shi2020pv}. The number of sampled keypoints $n$ is 2,048 for both Waymo and KITTI. In order to sample the keypoints effectively and accurately, we uses FPS on points within the range of top 1000 or 1500 initial proposals with a radius $r$ of 2.4.

\textbf{Waymo:} The voxel size is [0.1, 0.1, 0.15], and we focus on the input LiDAR point cloud range with [-75.2, 75.2], [-75.2, 75.2], and [-2, 4] meters in x, y, and z axis, respectively. The 2D ConvNets output size is $188 \times 188 \times 512$.

\textbf{KITTI:} The voxel size is [0.05, 0.05, 0.1], and we focus on the input LiDAR point cloud range with [0, 70.4], [-40, 40], and [-3, 1] meters in x, y, and z axis, respectively.  The 2D ConvNets output size is $200 \times 176 \times 512$.

\subsection{Detection Heads}
The RPN anchor size for each object category is set by computing the average of the corresponding objects from the annotated training set. The RoI-grid pooling module samples $ 6 \times 6 \times 6$ grid points within each initial 3D proposal to form a refined features. The number of surrounding points used to extract the grid point's feature, M, is 16.

During the training phase, 512 proposals are generated from RPN to R-CNN, where non-maximum suppression (NMS) with a threshold of 0.8 is applied to remove the overlapping proposals. In the validation phase, 100 proposals are fed into R-CNN, where the NMS threshold is 0.7.

\end{document}


\title{\textsc{M3DeTR}: Multi-representation, Multi-scale, Mutual-relation 3D Object Detection with Transformers\\Supplementary Material}

\maketitle

\section{Additional Ablation Study}
To further explore the effectiveness of M3DeTR, we conduct ablation on the Waymo Open Dataset. Specifically, we demonstrate the individual benefits of the multi-representation,
multi-scale, and mutual-relation layers of the M3 transformer. As shown in the Table~\ref{}, M3DeTR 

\section{More Implementation Details}
\subsection{RoIPool}
As pointed out in the Section 

\section{Additional Qualitative Results}
As shown in the Figure~\ref{}, we add some additional qualitative visualizations to further demonstrate the effectiveness of M3DeTR, compared to PV-RCNN~\cite{shi2020pv}.

\bibliographystyle{ACM-Reference-Format}
\bibliography{sample-base}